\documentclass{article}
\usepackage{spconf,amsmath,graphicx}
\usepackage{float} 
\usepackage{subfig}
\usepackage[export]{adjustbox} 
\usepackage{float} 
\usepackage{bbding}
\usepackage{booktabs}
\usepackage{multirow}
\usepackage{amssymb}
\usepackage{threeparttable}
\usepackage{stfloats}
\usepackage{xcolor}
\usepackage[colorlinks=true, linkcolor=red, citecolor=black, urlcolor=magenta]{hyperref}

\usepackage{pifont}
\usepackage{tabularx}
\title{NeRF-AD: Neural Radiance Field with Attention-based Disentanglement for Talking Face Synthesis}
\name{Chongke Bi\sthanks{Both authors contributed equally to the paper.}, Xiaoxing Liu\footnotemark[1], Zhilei Liu\sthanks{Corresponding author, Email: zhileiliu@tju.edu.cn.}}
\address{College of Intelligence and Computing, Tianjin University, Tianjin, China}
\begin{document}
\ninept
\maketitle
\begin{abstract}
Talking face synthesis driven by audio is one of the current research hotspots in the fields of multidimensional signal processing and multimedia. Neural Radiance Field (NeRF) has recently been brought to this research field in order to enhance the realism and 3D effect of the generated faces. 
However, most existing NeRF-based methods either burden NeRF with complex learning tasks while lacking methods for supervised multimodal feature fusion, or cannot precisely map audio to the facial region related to speech movements.
These reasons ultimately result in existing methods generating inaccurate lip shapes.
This paper moves a portion of NeRF learning tasks ahead and proposes a talking face synthesis method via NeRF with attention-based disentanglement (NeRF-AD). In particular, an Attention-based Disentanglement module is introduced to disentangle the face into Audio-face and Identity-face using speech-related facial action unit (AU) information. To precisely regulate how audio affects the talking face, we only fuse the Audio-face with audio feature. In addition, AU information is also utilized to supervise the fusion of these two modalities. Extensive qualitative and quantitative experiments demonstrate that our NeRF-AD outperforms state-of-the-art methods in generating realistic talking face videos, including image quality and lip synchronization. To view video results, please refer to \href{https://xiaoxingliu02.github.io/NeRF-AD/}{https://xiaoxingliu02.github.io/NeRF-AD/}.
\end{abstract}
\begin{keywords}
Talking face synthesis, neural radiance fields, facial disentanglement
\end{keywords}
\vspace{-6pt}
\section{Introduction}
\label{sec:intro}
\vspace{-3pt}
Synthesizing video of talking face from audio has a wide range of potential applications in virtual reality, entertainment, communication, and other areas. 
In existing research, talking face synthesis methods can be broadly classified into two categories: 2D methods~\cite{9878472, prajwal2020lip, 9496264} and 3D methods~\cite{zhang2021flow, peng2023selftalk}. 2D methods use image processing techniques to generate facial expressions and movements that match the speech. However, due to the fact that the face is essentially a 3D object, 2D methods always have limitations in generating facial texture and 3D realism \cite{9878472, prajwal2020lip, 9496264, zeng2020talking}. On the other hand, 3D methods construct a 3D parametric model of the face and animate it to match the audio. 
While these methods may produce more realistic facial videos, they face issues related to the complexity of the model and require a large dataset for training \cite{zhang2021flow, peng2023selftalk}.
Recently, some researchers have started to explore the use of Neural Radiance Field (NeRF) in talking face synthesis \cite{shen2022learning, guo2021ad, yao2022dfa, DBLP:conf/iclr/YeJ0LHZ23}. NeRF \cite{mildenhall2020nerf} is a recently proposed method for learning continuous volumetric representations of 3D scenes, which has shown promising results in synthesizing realistic images and videos using a limited dataset \cite{shen2022learning, guo2021ad, DBLP:conf/iclr/YeJ0LHZ23}.
\begin{figure*}[t]
\centering
  \includegraphics[scale=0.48]{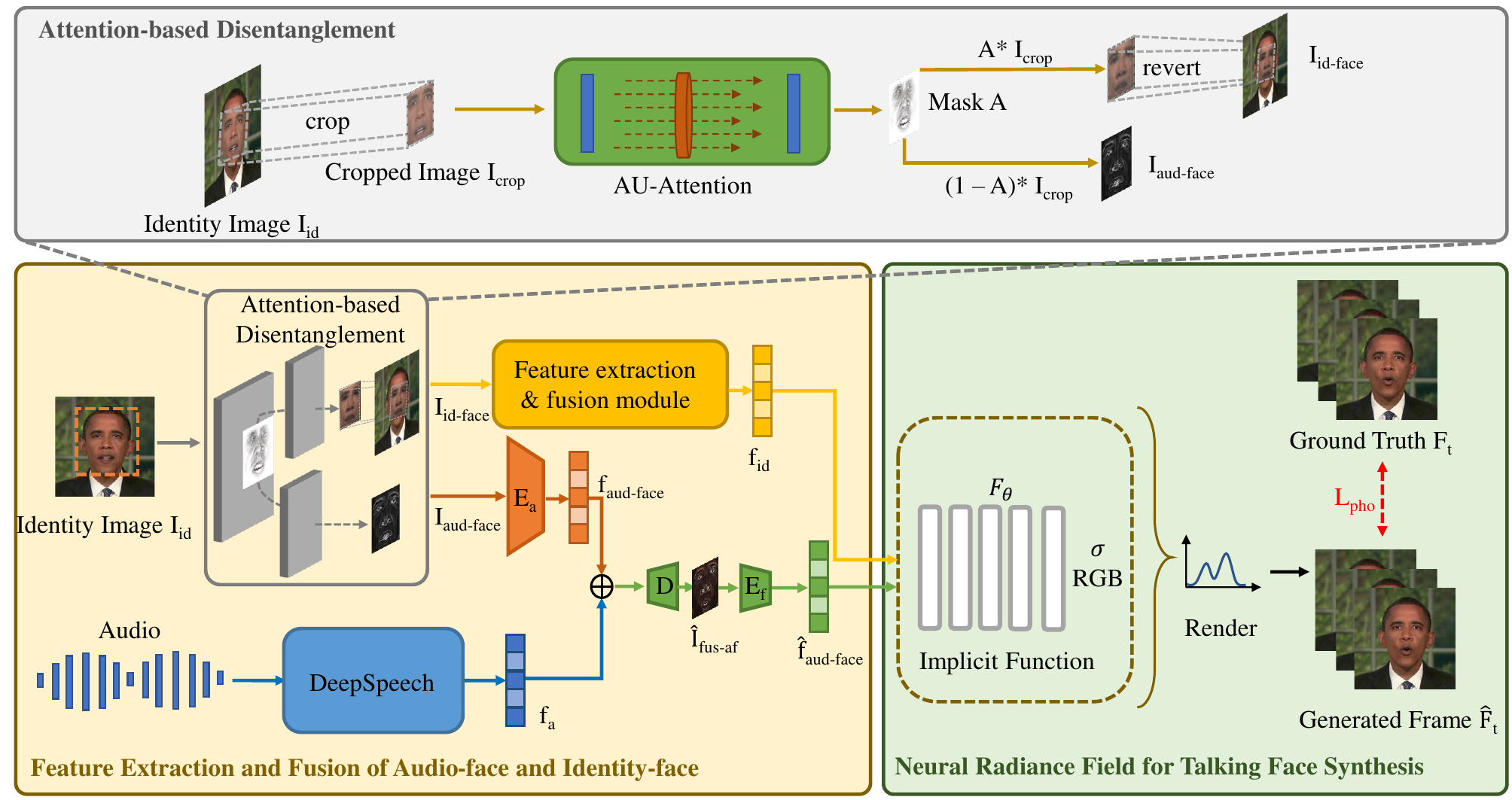}
  \vspace{-7pt}
  \caption{Overview of NeRF-AD. The gray area represents the Attention-based Disentanglement module proposed to disentangle the input talking face into Audio-face \(I_{aud-face}\) and Identity-face \(I_{id-face}\). Feature extraction \& fusion module is used to extract the identity feature \(f_{id}\) from \(I_{id-face}\). And encoder \(E_a\) is used to extract the Audio-face feature \(f_{aud-face}\). \(f_{aud-face}\) is then fused with audio feature \(f_a\) extracted from DeepSpeech module. Subsequently, decoder \(D\) generates feature map \(\hat{I}_{fus-af}\) supervised by AU information. After that, feature encoder \(E_f\) is used to generate the fused Audio-face feature vector \(\hat{f}_{aud-face}\). Finally, the conditional NeRF is presented to accurately render the talking face images with \(f_{id}\) and \(\hat{f}_{aud-face}\) as conditions.}
  \vspace{-12pt}
  \label{fig:teaser2}
\end{figure*}

However, recent methods that use NeRF in talking face synthesis either focus on the rendering efficiency of NeRF \cite{shen2022learning} or on the perception of the upper body \cite{DBLP:conf/iclr/YeJ0LHZ23, 10203662}. 
These methods \cite{shen2022learning, DBLP:conf/iclr/YeJ0LHZ23, 10203662} tend to use complete audio and face information to drive NeRF. In this scenario, the NeRF needs to accurately fuse the two modalities of features containing interference information and generate matching mouth shapes. 
Due to the complexity of the task and the lack of supervised feature fusion methods, the NeRF has to bear a significant learning burden. 
In addition, some other methods \cite{guo2021ad, yao2022dfa} also result in inaccurate mouth shapes due to the inability of audio to accurately affect the facial region related to speech movements.
To tackle the issues mentioned above, it is important to decompose the speech-related facial regions and use the decomposed precise features to drive the NeRF to synthesize the talking face images. 

Facial Action Coding System (FACS) is a comprehensive and objective facial action description system \cite{1978Facial, liu2020region}. It defines a set of basic facial action units (AUs), where each AU represents a basic facial muscle movement \cite{liu2020region}. 
In recent years, studies have successfully generated more vivid talking face videos using AU information \cite{chen2021talking}. 
This also indicates a strong correlation between AUs and speech movements. At the same time, attention mechanism allows the model to adaptively select specific parts \cite{pumarola2018ganimation} or features \cite{akram2023sargan} of an image. 
Hence, in this study, we utilize specific AUs related to speech movements to guide attention model in disentangling the face.

In this paper, we propose a talking face synthesis method via NeRF with attention-based disentanglement, which refers to as NeRF-AD.
At first, we design an Attention-based Disentanglement module that utilizes AUs to guide the attention model in generating masks for facial regions related to speech movements.
By utilizing these masks, we can effectively disentangle the input face into distinct components: the Audio-face and the Identity-face. 
The Audio-face represents the facial region related to speech movements, which is fused with audio feature, while the Identity-face represents the facial region related to speaker's identity. In this case, audio feature only affect the Audio-face, thus providing precise control over the generated face. Subsequently, we present a conditional NeRF to precisely render the talking face images with the fused Audio-face feature and Identity-face feature as conditions. In addition, we use AU loss to supervise the fusion process of Audio-face feature and audio feature, so that the two can be accurately fused. Throughout the process, we disperse the tasks of the NeRF and use different methods to supervise each task, making the NeRF more clear about what it needs to learn.
In summary, the contributions of this paper are summarized as follows:
\vspace{-3pt}
\begin{itemize}
\item To reduce the learning burden of NeRF and improve the accuracy of face rendering, we decompose the talking face and provide two decomposed precise conditions for the NeRF. 
\vspace{-3pt}
\item We propose an Attention-based Disentanglement module, allowing audio to be precisely fused with the facial region related to speech movements.
Meanwhile, we employ a range of methods to supervise the entire process.

\vspace{-3pt}
\item Extensive qualitative and quantitative experiments demonstrate that our proposed NeRF-AD outperforms the state-of-the-art in generating realistic and accurate talking faces.

\end{itemize}

\vspace{-13pt}
\section{Proposed method}
\label{sec:format}
\vspace{-5pt}
The proposed NeRF-AD is shown in Fig.\ref{fig:teaser2}. 
We divide the overall framework into three parts using big rectangles of three different colors (gray, pale yellow, pale green). The following three sections will respectively introduce these three parts.
\vspace{-11pt}

\subsection{Attention-based Disentanglement}
\label{Sec_Attention}
\vspace{-3pt}
We disentangle the image \(I_{id}\) into Audio-face and Identity-face using the Attention-based Disentanglement module (gray part of Fig.\ref{fig:teaser2}).
In this module, we crop the facial region of \(I_{id}\) and input it to the AU-Attention module to obtain mask \(A\), where \(A\) is defined as the weight of the facial region related to speech movements.

The detailed structure of the AU-Attention module is shown in Fig.\ref{fig:teaser3}, which consists of four branches. Mask Generator~\(G_M\) (composed of blue rectangle and green rectangle) is used to generate attention mask. Considering the lack of datasets for the facial disentanglement task,
we adopt an unsupervised approach and use Warping Generator \(G_W\) (composed of blue rectangle and yellow rectangle), AU Classifier \(C\) (pink rectangle) and Discriminator \(W_D\) (grey rectangle) to assist \(G_M\) in learning. After pre-training, we only retain \(G_M\) as AU-Attention module to generate attention mask \(A\).

Firstly, we randomly generate a target AU code from AUs related to speech movements. Then, we concatenate it with the image \(I_{crop}\) and input it into the Mask Generator \(G_M\) and Warping Generator \(G_W\), respectively.
And \(G_M\) will generate \(A\), specifying the weights of facial regions related to speech movements. \(G_W\) will generate a warped map \(I_W\) that is consistent with the AU code \(f_{AU_t}\). 
After that, we can get the final generated image \(I_G=A\cdot I_{crop}+(1-A)\cdot I_W\).

In this process, we use the following three loss functions:

\textbf{Speech-related code loss.}
We need to reduce the error in the speech-related information of the generated image \(I_G\). 
Specifically, we input \(I_G\) into AU Classifier \(C\) to generate AU label \(f_{AU_G}\) and reduce the distance between \(f_{AU_G}\) and \(f_{AU_t}\). At the same time, we also input the original image \(I_{crop}\) into \(C\) to generate AU label \(f_{AU_C}\), reducing the distance between \(f_{AU_C}\) and \(I_{crop}\)'s Ground Truth AU label \(f_{AU_O}\) to ensure the accuracy of \(C\). This process is as following:
\begin{equation}
    L_{AU_L}=||f_{AU_G}-f_{AU_t}||_2^2+||f_{AU_C}-f_{AU_O}||_2^2
\end{equation}

\textbf{Cycle identity loss.}
In the training process of this network, the speaker's identity information is easily biased, so the following loss function is used to ensure that the speaker's identity information will not be changed during the whole training process, and further improve the accuracy of the masks:
\begin{equation}
    L_{C}=||A_G\cdot I_{G}+(1-A_G)\cdot I_{G_W}-I_{crop}||_1
\end{equation}
where \(A_G=G_M(I_G,f_{AU_O})\) and \(I_{G_W}=G_W(I_G,f_{AU_O})\).
Specifically, we input the generated image $I_G$ into \(G_M\) and \(G_W\) to generate the original image and reduce loss between it and \(I_{crop}\).

\textbf{WGAN-GP loss.}
To enhance the realism of the generated images, the WGAN-GP \cite{gulrajani2017improved} is used to provide an image adversarial loss. The structure of the discriminator \(W_D\) is shown in the gray rectangle of Fig.\ref{fig:teaser3}. The specific loss function is as following:
\[L_W=\mathbb{E}_{I_G\sim \mathbb{P}_G}[W_D(I_G)]-\mathbb{E}_{I_{crop}\sim \mathbb{P}_{crop}}[W_D(I_{crop})]+\]
\begin{equation}
    \lambda\mathbb{E}_{\hat{I}\sim \mathbb{P}_{\hat{I}}}[(||\nabla W_D(\hat{I})||_2-1)^2] 
\end{equation}
where \(\hat{I}\) is random noise, \(\mathbb{P}_G\), \(\mathbb{P}_{crop}\) and \(\mathbb{P}_{\hat{I}}\) denote the distributions of generated images, crop images, and random noise respectively.

After obtaining the face mask $A$, we use \((1-A) \cdot I_{crop}\) to obtain the Audio-face \(I_{aud-face}\). We use \(A \cdot I_{crop}\) to obtain the face region that is not related to speech movements, after which we revert this part of the face region back onto the identity image to obtain the final Identity-face \(I_{id-face}\).

\vspace{-8pt}
\subsection{Feature Extraction and Fusion of Audio-face and Identity-face}
\label{Sec_AudioFace}
\vspace{-2pt}

For Identity-face \(I_{id-face}\), we use Feature extraction \& fusion module \cite{shen2022learning} to extract the identity feature \(f_{id}\). 
For Audio-face \(I_{aud-face}\), the encoder \(E_a\) is utilized to extract the Audio-face feature \(f_{aud-face}\). 
DeepSpeech \cite{hannun2014deep} is adopted to extract the audio feature \(f_a\). After that, we concatenate \(f_a\) and \(f_{aud-face}\) and input them into the decoder \(D\) to generate the fused Audio-face image \(\hat{I}_{fus-af}\). 
Here, we use AU loss and reconstruction loss to supervise the fusion process of multimodal features:
\begin{figure}[t]
  \centering
  \includegraphics[width=\linewidth]{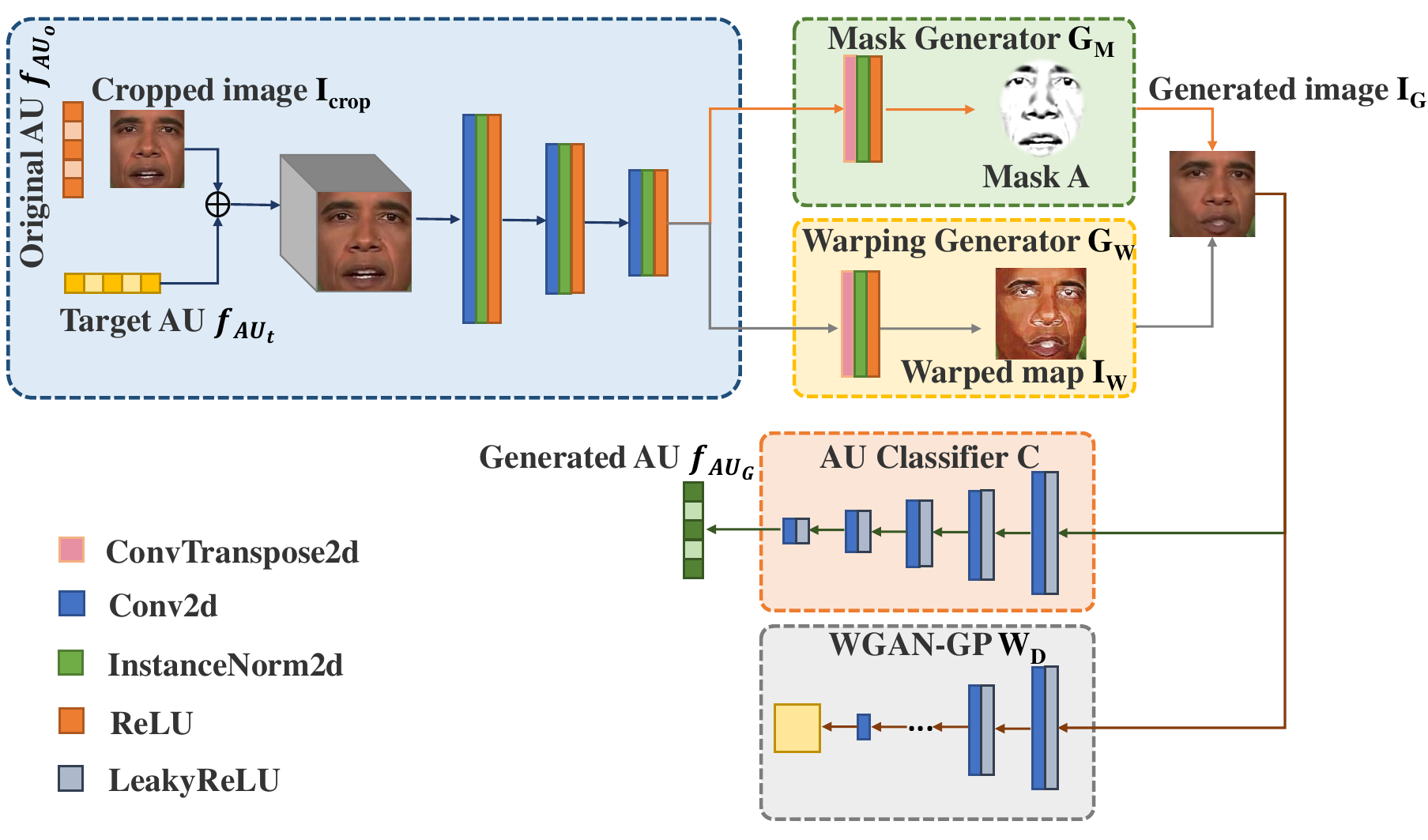}
  \caption{The framework of AU-Attention. }
  \label{fig:teaser3}
\vspace{-16pt}
\end{figure}

\textbf{AU loss.}
We use binary cross-entropy loss to compute the AU distance between \(\hat{I}_{fus-af}\) and ground truth \(I_{fus-af}\). \(I_{fus-af}\) is obtained by calculating \(F_t\) in Fig.\ref{fig:teaser2} with its corresponding mark. The specific loss function is shown as following:
\begin{equation}\label{bce}
    L_{bce}=-\frac{1}{n_{AU}}\sum_{i=1}^{n_{AU}}w_i[x_{i}log\hat{x}_{i}+(1-x_{i})log(1-\hat{x}_{i})]
\end{equation}
where \(\hat{x}_{i}\) and \(x_i\) denote the i-th AU label of \(\hat{I}_{fus-af}\) and the i-th AU label of \(I_{fus-af}\) respectively. 
The AU labels are extracted using OpenFace \cite{baltrusaitis2018openface}. 
To reduce the impact of the correlation between AUs on model training, we add weights \(w_i=\frac{\frac{1}{r_i}n_{AU}}{\sum_{i=1}^{n_{AU}}\frac{1}{r_i}}\) to Eq.~\eqref{bce}, where \(r_i\) is the probability of the $i$-th AU occurring in the dataset. 
Meanwhile, since some AUs appear rarely in the training set and can affect the model's prediction results, we introduce a weighted multi-label Dice coefficient loss \cite{milletari2016v}.
\begin{equation}
    L_{dice}=\frac{1}{n_{AU}}\sum_{i=0}^{n_{AU}}w_i[1-\frac{2x_i\hat{x}_{i}+\varepsilon}{x_i^2+\hat{x}_{i}^2+\varepsilon}]
\end{equation}

Finally, our total AU loss is \(L_{AU}=L_{bce}+L_{dice}\).

\textbf{Reconstruction loss.}
We use reconstruction loss to reduce the pixel difference between \(\hat{I}_{fus-af}\) and \(I_{fus-af}\): 
\begin{equation}
    L_{rec}=||I_{fus-af}-\hat{I}_{fus-af}||_1
\end{equation}

After that, we input \(\hat{I}_{fus-af}\) into the feature encoder \(E_f\) to obtain the fused Audio-face feature vector \(\hat{f}_{aud-face}\).

\begin{table*}[]
\resizebox{\textwidth}{!}{
\begin{tabular}{c|cccccc|cccccc|c}
\toprule
\multirow{2}{*}{Method} & \multicolumn{6}{c|}{TestA} & \multicolumn{6}{c|}{TestB} & \multirow{2}{*}{\textcolor{blue}{Ne}} \\ \cline{2-13}
                & PSNR↑& SSIM↑  & LPIPS↓ & AU Acc↑ & LMD-79↓  & SyncNet↑   & PSNR↑& SSIM↑   & LPIPS↓  & AU Acc↑        & LMD-79↓  & SyncNet↑   &      \\ \midrule \midrule
    Ground-truth &-    & -      & -      & -       & -        & 3.176      & -     & -                   &-        &-               &-         &2.705       & -     \\
    Wav2Lip 2020\cite{prajwal2020lip}&31.980  &  0.828 &0.068    & 86.3\%    & 4.069      & \textbf{2.381}  &27.293  &0.695 &0.206   &77.0\%  &4.934  &\textbf{1.987} & \ding{53} \\
    Sefik et al. 2022\cite{9496264} &31.807  &0.833 &0.055  &83.3\% & 4.465  &1.835 &26.907  & 0.709 &0.193 &74.8\% &5.170 &1.673 & \ding{53} \\
    NeRF 2020\cite{mildenhall2020nerf} & 32.110  & 0.828  & 0.085 & 79.6\% & 4.942 & 0.961  & 27.886   & 0.625  & 0.264  & 69.8\% & 6.257  & 1.074  & \ding{51}                \\
    AD-NeRF 2021\cite{guo2021ad} & 35.722 & 0.899 & 0.043 & 86.3\% & 4.486 & 1.862  & 29.409 & 0.707  & 0.188  & 76.5\%  & 4.957  & 1.607    & \ding{51}               \\
    DFRF 2022\cite{shen2022learning}  & \textbf{36.016} & 0.897 & 0.042  & 85.9\%   & 4.469  & 1.968  & 28.836 & 0.702  & 0.180 & 77.6\%  & 4.923 & 1.801  & \ding{51} \\ 
    GeneFace 2023\cite{DBLP:conf/iclr/YeJ0LHZ23}  & 34.837 & 0.897 & 0.047  & 85.2\%   & 4.342  & 2.131  & 29.040 & 0.689  & 0.186 & 77.5\%  & 5.061 & 1.848  & \ding{51} \\ \midrule
    \textbf{Ours} & 35.715 & \textbf{0.901} & \textbf{0.039} & \textbf{87.5\%} & \textbf{3.950} & 2.303 & \textbf{29.557} & \textbf{0.712} & \textbf{0.180} & \textbf{80.8\%} & \textbf{4.725} & 1.937& \ding{51}  \\ \bottomrule
\end{tabular}}
\vspace{-6pt}
\caption{Quantitative results compared with other methods. Best results are in \textbf{bold}. \textcolor{blue}{"Ne"} indicates whether this method is based on NeRF.}
\vspace{-10pt}
\label{tab:t1}
\end{table*}

\subsection{Neural Radiance Field for Talking Face Synthesis}
\label{Sec_NeRF}
\vspace{-3pt}
We present a conditional NeRF to synthesize the talking face images with \(\hat{f}_{aud-face}\) and \(f_{id}\) as conditions. 
Then, we use these conditions along with 3D position \(l\) and view direction \(d\) as the input to the implicit function \(F_\theta\). After that, the NeRF, consisting of multi-layer perceptrons, can estimate the color \(c\) and density \(\sigma\) of each 3D point on every ray. 
The entire implicit function can be formulated as follows:
\begin{equation}
    F_\theta:(l,d,f_{id},\hat{f}_{aud-face})\longrightarrow(c,\sigma)
\end{equation}

To calculate the color \(C_n\) of each pixel on the output image, we perform volume rendering by integrating the colors and densities of all 3D points along each ray \(r\). This process can be represented as:
\begin{equation}
    C_n(r;\theta,f_{id},\hat{f}_{aud-face})=\int_{t_n}^{t_f}\sigma(t)\cdot c(t)\cdot T(t)dt
\end{equation}
where \(t_n\) and \(t_f\) denote the near and far bound of a camera ray respectively. \(T(t)=exp(-\int_{t_n}^{t}\sigma(r(\tau))d\tau)\) denotes the integral transmittance along a camera ray from near \(t_n\) bound to $t$. In this process, we use the loss function \(L_{pho}\) to minimize the photo-metric reconstruction error between the generated color \(C_n\) and the ground truth color \(C_g\).
\begin{equation}
    L_{pho}=\sum_{r\in \mathcal{R}}||C_n(r;\theta,f_{id},\hat{f}_{aud-face})-C_g||^2
\end{equation}
where \(\mathcal{R}\) denotes the set of rays.

\vspace{-10pt}
\section{EXPERIMENTS}
\label{sec:pagestyle}
\vspace{-5pt}

\textbf{Dataset.}
To compare with SOTA methods, we obtain the dataset from publicly released video sets \cite{shen2022learning, guo2021ad, DBLP:conf/iclr/YeJ0LHZ23}.
The dataset includes seven videos of different individuals, each with a frame rate of 25fps and an average of about 7,000 frames. 
Similar to the training process in \cite{shen2022learning}, we select five videos to train NeRF-AD. The remaining two videos is segmented. A random portion from these two videos is used as the fine-tuning training set, and the other portion as the test set. We strictly ensure that the training set and test set do not overlap.

\textbf{Evaluation Metrics.}
We use the metrics of PSNR, SSIM \cite{wang2004image}, and LPIPS \cite{zhang2018unreasonable} to evaluate the quality of images. To evaluate lip synchronization, we use AU Acc \cite{chen2021talking}, landmark distance (LMD) \cite{chen2018lip}, and SyncNet confidence value \cite{prajwal2020lip}. It is worth noting that to more accurately evaluate the lip shape, we improve the LMD metric by increasing the number of calculated lip landmarks from 20 to 79 using MediaPipe \cite{kartynnik2019real}. We denote this metric as LMD-79.

\vspace{-9pt}
\subsection{Quantitative Results}
\vspace{-3pt}
The quantitative comparison results between our NeRF-AD and other methods are shown in Tab.\ref{tab:t1}.

\textbf{Comparison with NeRF-based Methods} (\textcolor{blue}{"Ne"} in Tab.\ref{tab:t1} is \ding{51}).
First, we compare our NeRF-AD with state-of-the-art NeRF-based methods, including GeneFace \cite{DBLP:conf/iclr/YeJ0LHZ23}, DFRF~\cite{shen2022learning}, AD-NeRF \cite{guo2021ad} and origional NeRF~\cite{mildenhall2020nerf}. 
Since PSNR can sometimes give higher values for blurry images \cite{park2021nerfies}, we recommend readers to pay more attention to the other five metrics. From Tab.\ref{tab:t1}, it can be seen that compared with NeRF-based methods, our NeRF-AD performs the best on the other five metrics except for PSNR. 
In terms of lip synchronization metrics, compared with GeneFace, our method increases AU Acc↑ by 2.3\% and reduces LMD-79↓ by 0.39 on Test A. 

\textbf{Comparison with Non-NeRF-based Methods} (\textcolor{blue}{"Ne"} in Tab.\ref{tab:t1} is \ding{53}).
To conduct a more comprehensive evaluation, we also compare our method with state-of-the-art non-NeRF-based methods, including Wav2Lip \cite{prajwal2020lip} and the method proposed by Sefik et al. \cite{9496264}. Compared with them, our method demonstrates significant advancements in image quality. Regarding lip synchronization, since Wav2Lip is jointly trained with a sync-expert on a large dataset, it slightly outperforms NeRF-AD on the SyncNet metric. However, our method performs best on the AU Acc and LMD-79 metrics.
\vspace{-12pt}

\subsection{Qualitative Results}
\begin{figure}[t]
  \centering
  \includegraphics[width=\linewidth]{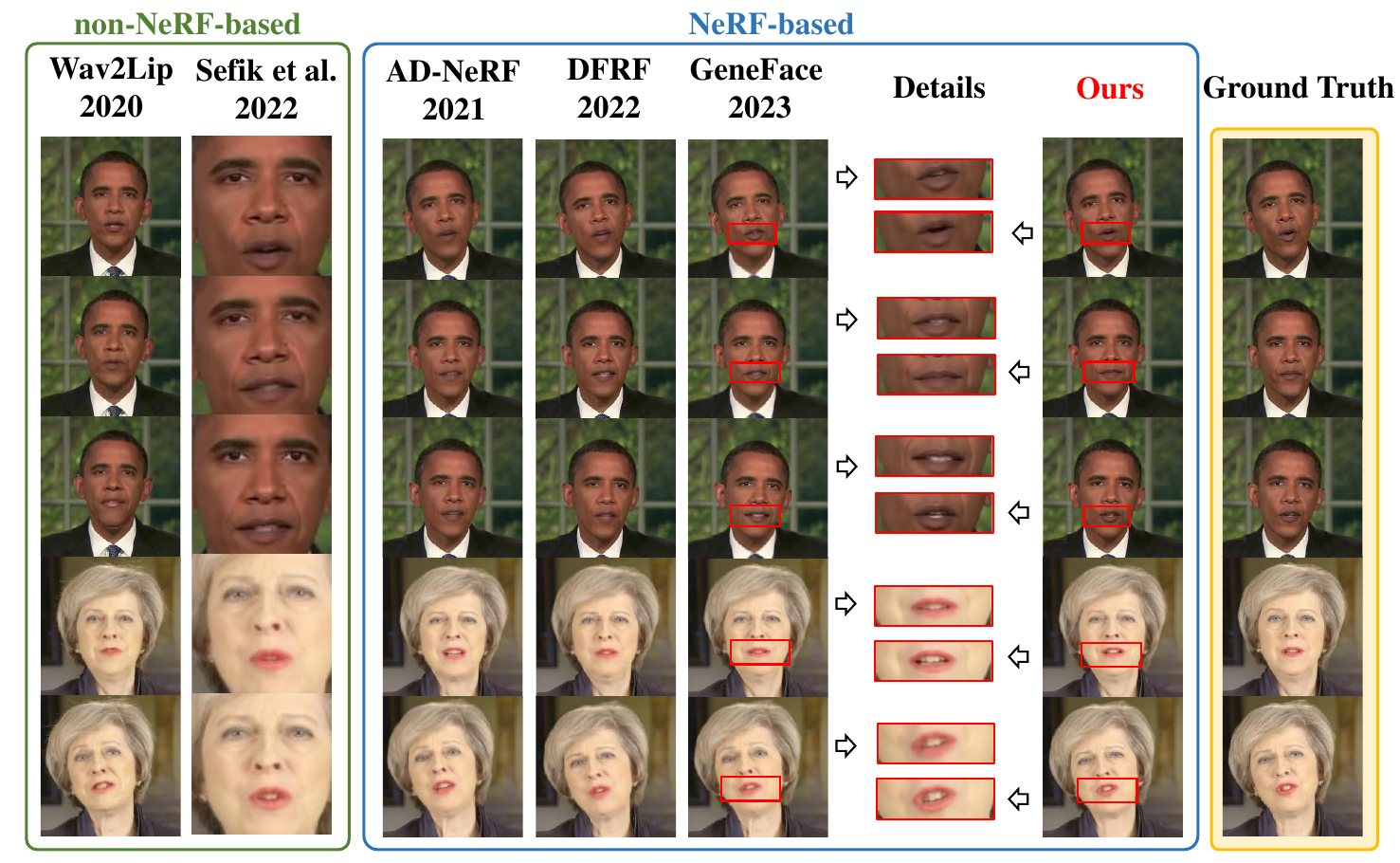}
  \vspace{-15pt}
  \caption{Visual results of the comparative experiments.}
  \label{fig:teaser5}
\vspace{-19pt}
\end{figure}

\vspace{-3pt}
The qualitative comparison results with other methods are shown in Fig.\ref{fig:teaser5}. From the figure, it is evident that our method generates lip movements that are closest to the ground truth. First, the lip movements generated by other NeRF-based methods are more prone to deviations and incorrect. 
This is because other NeRF-based methods burden NeRF with complex learning tasks and they cannot precisely map audio to the facial region related to speech movements. 
Second, due to Wav2Lip solely focusing on lip synchronization, the generated images have lower image quality. And the method proposed by Sefik et al. cannot generate natural head poses. So, whether in comparison with NeRF-based methods or non-NeRF-based methods, our method generates the best visual results.

\vspace{-12pt}
\subsection{Ablation Studies}
\vspace{-3pt}

\textbf{Learning Efficiency of NeRF.}
In this section, we compare our model with NeRF-based models at different iterations. The quantitative results are shown in Tab.\ref{tab:t3}. From the table, it can be seen that our model outperforms the other models in both image quality and lip synchronization when these models are iterated for only 4k and 10k. When the number of iterations is 10k, compared with GeneFace, our model has a 2.5\% higher AU Acc↑ and is 0.26 lower than LMD-79↓.
This indicates that our model has the highest learning efficiency, enabling the generation of the most accurate faces within the same limited number of iterations.
\begin{table}[h!t]
\tabcolsep=0.27cm
\resizebox{\linewidth}{!}{
\begin{tabular}{ccccccc}
\toprule
\multicolumn{1}{c|}{Method}   & PSNR↑  & SSIM↑ & LPIPS↓ & AU Acc↑ & LMD-79↓  & SyncNet↑ \\ 
\midrule \midrule
\multicolumn{1}{c|}{Ground-truth}   & -      & -     & -      & -       & -     &     2.705     \\ 
\midrule
\multicolumn{7}{c}{4k}  \\
\midrule
\multicolumn{1}{c|}{NeRF\cite{mildenhall2020nerf}} &  27.460      &  0.613     &     0.319   &   67.8\%      &  8.402     &   0.825       \\
\multicolumn{1}{c|}{AD-NeRF\cite{guo2021ad}} &  28.244     &  0.635     &     0.304   &   73.3\%      &  7.336     &   1.037       \\
\multicolumn{1}{c|}{DFRF\cite{shen2022learning}} & 27.687 & 0.669 & 0.246  & 73.7\%   & 6.842 & 1.115    \\
\multicolumn{1}{c|}{GeneFace\cite{DBLP:conf/iclr/YeJ0LHZ23}} & 27.642 & 0.627 & 0.262  & 74.4\%   & \textbf{6.836} & 0.932    \\
\multicolumn{1}{c|}{Ours} & \textbf{28.512} & \textbf{0.675} & \textbf{0.227}  & \textbf{77.3\%}   & 6.938 & \textbf{1.200}    \\ 
\midrule
\multicolumn{7}{c}{10k} \\
\midrule
\multicolumn{1}{c|}{NeRF\cite{mildenhall2020nerf}} &  27.836      &   0.647    &   0.279     &   69.4\%      &   8.290    &    0.979      \\
\multicolumn{1}{c|}{AD-NeRF\cite{guo2021ad}} &  28.793     &  0.662     &     0.268   &   75.3\%      &  6.520     &   1.116       \\
\multicolumn{1}{c|}{DFRF\cite{shen2022learning}} & 28.429 & 0.686 & 0.237  & 77.3\%   & 6.453 & 1.212    \\
\multicolumn{1}{c|}{GeneFace\cite{DBLP:conf/iclr/YeJ0LHZ23}} & 28.148 & 0.660 & 0.240  & 77.1\%   & 6.569 & 1.062    \\
\multicolumn{1}{c|}{Ours} & \textbf{29.059} & \textbf{0.689} & \textbf{0.220}  & \textbf{79.6\%}   & \textbf{6.309} & \textbf{1.400}    \\ 
\bottomrule
\end{tabular}}
\vspace{-8pt}
\caption{Method comparisons at different training iterations.}
\vspace{-17pt}
\label{tab:t3}
\end{table}

\textbf{Effect of Different Components of NeRF-AD.}
To quantify the effect of the main components of our proposed model, we conduct ablation studies, and the results are shown in Tab.\ref{tab:t4}. The entries "w/o AU" and "w/o att-d" indicate the experimental results after removing the AU Loss and the Attention-based Disentanglement module, respectively. 
From the table, we can see that removing these components step by step results in a decrease in the quality of the generated images and the lip synchronization. Especially lip synchronization shows a significant decrease.
These observations indicate that: 1) Without the supervision of AU Loss, the fusion of \(f_a\) and \(f_{aud-face}\) in Section~\ref{Sec_AudioFace} will become inaccurate. 
2) Without the Attention-based Disentanglement module, the input complete audio and identity images fed into the NeRF contain a lot of interfering information, which increases the complexity of the learning task for the NeRF, resulting in a decline in lip synchronization.
\begin{table}[h!t]
\tabcolsep=0.27cm
\resizebox{\linewidth}{!}{
\begin{tabular}{ccccccc}
\toprule
\multicolumn{1}{c|}{Method}                                & PSNR↑           & SSIM↑          & LPIPS↓         & AU Acc↑        & LMD-79↓           & SyncNet↑       \\ \midrule \midrule
\multicolumn{7}{c}{TestA} \\ \midrule
\multicolumn{1}{c|}{Ground-truth}  & -    & -              & -              & -              & -              & 3.176          \\ 
\multicolumn{1}{c|}{w/o att-d}       & \textbf{35.940} & 0.899          & 0.042          & 80.8\%          & 4.460          & 1.970          \\
\multicolumn{1}{c|}{w/o AU}             & 35.641          & 0.900          & \textbf{0.039} & 82.4\%          & 4.209          & 2.199          \\
\multicolumn{1}{c|}{Ours}          & 35.715          & \textbf{0.901} & \textbf{0.039} & \textbf{87.5\%} & \textbf{3.950} & \textbf{2.303} \\ \midrule
\multicolumn{7}{c}{TestB} \\ \midrule
\multicolumn{1}{c|}{Ground-truth}   & -               & -              & -              & -              & -              & 2.705          \\ 
\multicolumn{1}{c|}{w/o att-d}            & 29.288          & 0.698          & 0.209          & 78.0\%          & 5.175          & 1.646          \\
\multicolumn{1}{c|}{w/o AU}             & 29.409          & 0.707          & 0.206          & 80.4\%          & 4.957          & 1.850          \\
\multicolumn{1}{c|}{Ours}                & \textbf{29.557} & \textbf{0.712} & \textbf{0.180} & \textbf{80.8\%} & \textbf{4.725} & \textbf{1.937} \\ \bottomrule
\end{tabular}}
\vspace{-8pt}
\caption{Quantitative results of the ablation study. }
\label{tab:t4}
\vspace{-22pt}
\end{table}

\section{CONCLUSION}
\label{sec:typestyle}
\vspace{-4.6pt}
In this paper, we propose NeRF-AD to address the high complexity of the NeRF learning task in talking face synthesis. NeRF-AD uses Attention-based Disentanglement module to disentangle the talking face into Audio-face and Identity-face. We only fuse the Audio-face feature and audio feature to accurately control the impact of audio on the talking face. Moreover, AU information is used to control the precise fusion of different modal features.
Results of extensive qualitative and quantitative experiments demonstrate that NeRF-AD outperforms other state-of-the-art methods in both image quality and lip synchronization.

\textbf{Acknowledgement.} This work was supported by National Natural Science Foundation of China under Grant No. 62172294.



\bibliographystyle{IEEEbib}
\bibliography{strings,refs}

\end{document}